\documentclass[sigconf]{acmart}
\AtBeginDocument{%
  }

\copyrightyear{2026}
\acmYear{2026}
\setcopyright{cc}
\setcctype{by}
\acmConference[CIKM '26]{Proceedings of the 35th ACM International Conference on Information and Knowledge Management}{November 07--11, 2026}{Rome, Italy}
\acmBooktitle{Proceedings of the 35th ACM International Conference on Information and Knowledge Management (CIKM '26), November 07--11, 2026, Rome, Italy}
\acmDOI{10.1145/3799682.3840714}
\acmISBN{979-8-4007-2539-5/2026/11}

\usepackage{multirow}

\usepackage{booktabs}
\usepackage[utf8]{inputenc}
\usepackage{xcolor}
\usepackage{CJKutf8} % 引入 CJK 中文包
\usepackage{tcolorbox}
\tcbuselibrary{skins, breakable}

\definecolor{lightgray}{RGB}{240,240,240}
\definecolor{lightorange}{RGB}{255,243,224}
\definecolor{lightblue}{RGB}{224,239,255}
\definecolor{promptgreen}{RGB}{0,128,0}
\definecolor{promptblue}{RGB}{0,0,160}
\definecolor{promptred}{RGB}{180,0,0}
\definecolor{defaultblack}{RGB}{0,0,0}

\newtcolorbox{promptbox}[2][]{
    enhanced,
    breakable,
    colback=#2,
    colframe=#2,
    arc=5pt,
    boxrule=0.5pt,
    left=10pt,
    right=10pt,
    top=8pt,
    bottom=8pt,
    fontupper=\normalfont\small,
    before upper={\parindent0pt},
    #1
}
\begin{document}

\title{Sandwich Reasoning: An Answer-Reasoning-Answer Approach for Low-Latency Query Correction}

\author{Chen Zhang}
\authornote{Chen Zhang and Kepu Zhang contributed equally to this work.}
\affiliation{%
  \department{Gaoling School of Artificial Intelligence}
  \institution{Renmin University of China}
  \city{Beijing}
  \country{China}
}
\email{2025104112@ruc.edu.cn}

\author{Kepu Zhang}
\authornotemark[1]
\affiliation{%
 \department{Gaoling School of Artificial Intelligence}
  \institution{Renmin University of China}
  \city{Beijing}
  \country{China}
}
\email{kepuzhang@ruc.edu.cn}

\author{Jiatong Zhang}
\affiliation{%
 \department{Gaoling School of Artificial Intelligence}
  \institution{Renmin University of China}
  \city{Beijing}
  \country{China}
}
\email{2023200400@ruc.edu.cn}

\author{Xiao Zhang}
% \correspondingauthor
\authornote{Xiao Zhang is the corresponding author (e-mail: zhangx89@ruc.edu.cn). 
% Work partially done at Engineering Research Center of Next-Generation Intelligent Search and Recommendation, Ministry of Education. 
}
\affiliation{%
 \department{Gaoling School of Artificial Intelligence}
  \institution{Renmin University of China}
  \city{Beijing}
  \country{China}
}
\email{zhangx89@ruc.edu.cn}

\author{Jun Xu}
\affiliation{%
 \department{Gaoling School of Artificial Intelligence}
  \institution{Renmin University of China}
  \city{Beijing}
  \country{China}
}
\email{junxu@ruc.edu.cn}

\renewcommand{\shortauthors}{Chen Zhang, Kepu Zhang, Jiatong Zhang, Xiao Zhang, and Jun Xu}
\begin{abstract}
Query correction is a critical entry point in modern search pipelines, demanding high accuracy strictly within stringent
latency constraints. 
Chain-of-Thought (CoT) reasoning improves accuracy but introduces excessive latency that limits real-time deployment. A natural alternative is to output an answer before reasoning to reduce latency; 
however, under autoregressive decoding, such an answer-first paradigm suffers from a \emph{decoupling
problem}: the early answer is generated independently of the subsequent
reasoning, and therefore cannot benefit from it. To address this issue, we propose Sandwich Reasoning (SandwichR), a novel approach that explicitly aligns a fast initial answer with post-hoc reasoning, enabling low-latency query correction without sacrificing reasoning-aware accuracy.
SandwichR follows an ``Answer–Reasoning–Answer'' paradigm, producing an initial  correction, an explicit reasoning process, and a final refined correction.
To align the initial answer with post-reasoning insights, we design a consistency-aware reinforcement learning (RL) strategy: a dedicated consistency reward enforces alignment between the initial and final corrections, while margin-based rejection sampling prioritizes borderline samples where reasoning drives the most impactful corrective gains. Additionally, unlike existing datasets that primarily focus on simple spelling errors, we construct a high-quality query correction dataset tailored to semantically sensitive errors.
Experiments on three real-world domains show that
SandwichR delivers the best accuracy-latency trade-off among evaluated
methods at comparable model scale, achieving accuracy on par with
standard CoT while reducing inference latency by 40--70\%, and offering
a promising solution for latency-sensitive applications.
\end{abstract}





\maketitle
\begin{figure}[htbp]  % 浮动位置参数：优先此处(here)/顶部(top)/底部(bottom)/单独页(page)
    \centering        % 图片居中（必须放在figure环境内）
    % 插入图片：width设置宽度（推荐用相对宽度，如0.8\textwidth），{}内是图片路径
    \includegraphics[width=0.45\textwidth]{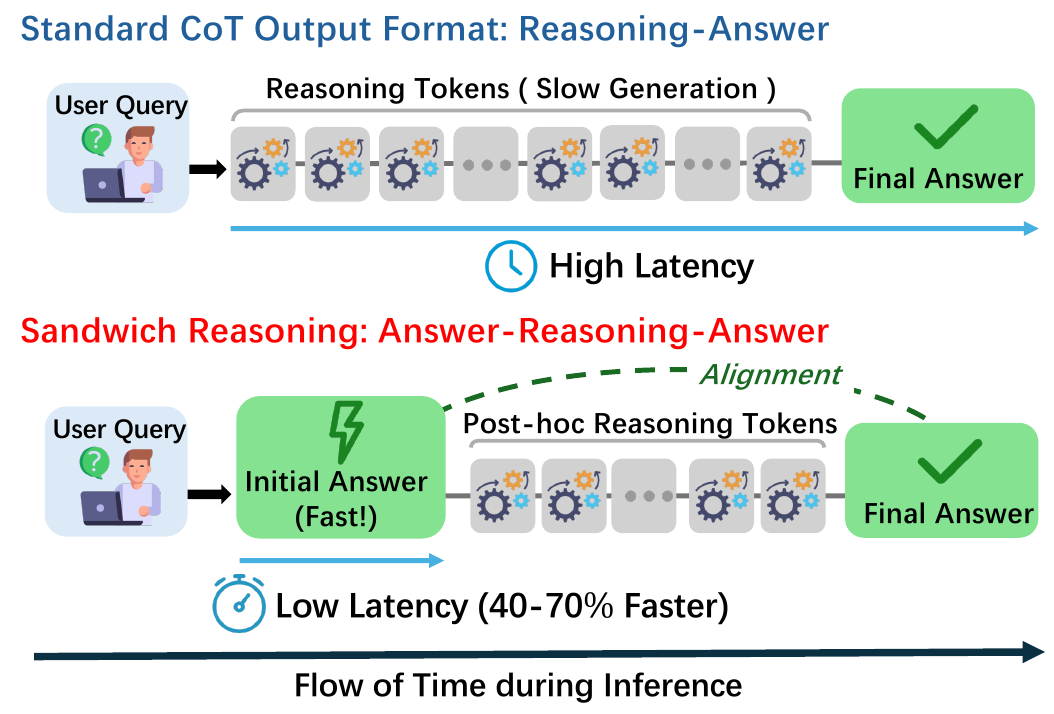}
    % 图片描述（自动在图片下方，会自动生成“图1：xxx”编号）
    \caption{Comparison of reasoning paradigms: traditional Chain-of-Thought (CoT) reasoning vs the proposed sandwich reasoning in this paper.}
    \label{fig:inference_comparison}  % 可选：图片标签，用于交叉引用（如\ref{fig:labelname}）
\end{figure}

\section{Introduction}

Query correction~\cite{ye2023improving,pande2022learning,zhang2025trigger3} serves as the first line of defense against noisy user inputs in modern Information Retrieval (IR) systems.
% Query correction~\cite{ye2023improving,pande2022learning,zhang2025trigger3} serves as the first line of defense in modern Information Retrieval (IR) systems. User queries often contain various noise, such as phonetic errors, typos, and semantic ambiguities, which can significantly degrade retrieval relevance. 
While Large Language Models (LLMs) have demonstrated remarkable capabilities in text processing, deploying them for real-time query correction faces a critical dilemma: the trade-off between accuracy and inference latency.

% The Efficiency-Accuracy Dilemma
% 推理能提升效果，带来效率问题，画图
% 引入问题，效率的时间放上，多少ms
% 直接生成，推理，时间，超过多少s
Chain-of-Thought (CoT) reasoning has demonstrated its effectiveness across a wide range of tasks~\cite{wang2024chain,han2024token,wang2023can}; it can enhance query correction accuracy by allowing the model to ``think'' before it outputs the corrected query. However, this \emph{reasoning-first} paradigm incurs high computational costs and unacceptable latency for online search scenarios.

% Conversely, Direct Generation methods are efficient but often lack robustness when dealing with complex, context-dependent errors. In cognitive science terms, we desire a system that possesses the speed of System 1 (intuition) while retaining the rigorous logic of System 2 (reasoning).
% \zc{
An intuitive solution is to reverse the CoT order: generate an answer first, then reason about it~\cite{dong2025taosr1thinkingmodelecommerce}. This promises the efficiency of a direct response. However, within a standard autoregressive model, this simple \emph{answering-first} approach suffers from a decoupling problem: the initial answer is generated in isolation, blind to the reasoning that follows, thus gaining no actual benefit from it. In this setting, the reasoning trace may become only a post-hoc
explanation of an already generated correction. Although such a
format reduces user-facing latency, it provides no explicit mecha-
nism for improving the early answer itself. 
% }

% sandwich-qc
% 先回答再推理有什么问题，自回归来说就独立了
% 先ans在推理，为什么不适合
% 再我们的方法

To bridge this gap, we introduce a novel answering-first approach for query correction, \textbf{Sandwich Reasoning}, which we refer to as \textbf{SandwichR}. Unlike standard CoT,  as shown in Figure~\ref{fig:inference_comparison}, our SandwichR outputs a sequence in an ``Answer-Reasoning-Answer'' format: an initial correction, followed by a reasoning trajectory, and a final correction. Unlike a plain Answer--Reasoning format, SandwichR adds a second correction after the reasoning trace during training. This final correction provides an explicit target for aligning the early answer with the model's own reasoning-refined output. At deployment, this structure allows the downstream search engine to utilize the initial correction for low-latency retrieval.
% while asynchronously using the reasoning and $C_{final}$ for verification or re-ranking.

%However, simply training a model to output this format often leads to a disconnect where initial correction remains prone to errors, and the subsequent reasoning becomes mere post-hoc rationalization. To address this, 
To ensure the first-generated query correction correlates with the subsequent reasoning process, 
we propose a consistency-aware reinforcement learning (RL) strategy during training. By rewarding consistency between the initial and final corrections during training, we effectively distill the model's own reasoning capabilities back into its immediate intuition, enabling the initial correction to achieve accuracy comparable to CoT even without explicit reasoning steps during inference. To stabilize RL training, we employ a rejection sampling strategy, filtering for samples where the model shows potential for self-correction. 
% \zc{
%We illustrate the inference process of the traditional Reasoning-Answer approach versus our proposed SandwichR framework in Figure~\ref{fig:inference_comparison}.
% } 
Furthermore, existing query correction datasets primarily focus on simple spelling or phonetic errors~\cite{jiang2022mcscsetspecialistannotateddatasetmedicaldomain,10.1145/3616855.3635847,ye-etal-2025-qspell}, lacking benchmarks for complex, context-dependent errors requiring reasoning. To address this, we construct a high-quality, controlled benchmark derived from real search queries by injecting representative error patterns. 
%Our experiments demonstrate that SandwichR achieves SOTA performance while maintaining the low latency.
Our contributions are as follows:

\begin{itemize}
\item We propose SandwichR, an Answer–Reasoning–Answer framework that returns an initial correction before generating reasoning tokens, thereby reducing user-facing latency while retaining a reasoning-refined training signal.
% We propose SandwichR, a ``Answer-Reasoning-Answer'' framework that effectively mitigates the dilemma between accuracy and latency by decoupling the reasoning process from the initial response for latency.

% We propose SandwichR, an “Answer-Reasoning-Answer” framework that addresses this dilemma by providing an initial answer to ensure low latency, while aligning it with subsequent reasoning to enhance its accuracy.

\item 
We design a consistency-aware RL strategy with margin-based rejection sampling that aligns the model's fast first-pass correction with its slower reasoning-refined correction, effectively distilling CoT correction capability into the initial answer.
\item We construct a realistic query correction benchmark based on a retrieval dataset and demonstrate that our method 
% achieves SOTA performance while delivering a remarkable 40--70\% speedup over the standard CoT approach, offering a highly promising solution for latency-sensitive applications.

achieves the best performance among trained models at the same scale, while delivering a 40--70\% speedup over the standard CoT approach, offering a promising solution for latency-sensitive applications.
\end{itemize}

\section{Related Work}
{\bf Query Correction (QC)} serves as a pivotal component in the initial stage of search engine pipelines~\cite{ye2023improving,gao2010large}, directly influencing retrieval recall and user satisfaction. Traditional approaches typically formulate query correction as a sequence-to-sequence translation task, evolving from statistical language models to Pre-trained Language Models (PLMs) such as BART~\cite{shao2024cpt}, which map noisy queries to their corrected forms based on contextual likelihood. With the emergence of Large Language Models (LLMs), recent studies have explored leveraging the extensive world knowledge of LLMs for correction via few-shot prompting~\cite{davis2024prompting,li2023effectiveness} or supervised fine-tuning~\cite{fan2023grammargpt}. While LLMs demonstrate superior semantic understanding compared to smaller models, they often suffer from over-correction—erroneously altering correct named entities or shifting the user's original intent. Crucially, most existing works treat correction as an immediate generation task. There is a notable scarcity of research investigating the application of explicit reasoning mechanisms to QC, which limits the model's robustness when facing ambiguous or complex errors. Furthermore, existing QC datasets and benchmarks~\cite{jiang2022mcscsetspecialistannotateddatasetmedicaldomain, 10.1145/3616855.3635847, ye-etal-2025-qspell} predominantly focus on simple spelling mistakes or phonetic typos, motivating our specialized benchmark designed for complex, context-dependent errors that necessitate explicit reasoning.  

{\bf Reasoning Large Language Models (LLMs)} have advanced significantly via Chain-of-Thought (CoT) reasoning, improving performance in complex domains such as mathematics and logic~\cite{guo2025deepseek,jaech2024openai}. To further optimize reasoning capabilities, researchers have focused on both data-centric approaches—selecting high-quality reasoning trajectories~\cite{ye2025limo}—and algorithmic innovations, such as designing granular process rewards or employing Reinforcement Learning to align model behaviors~\cite{aggarwal2025l1,han2024token,zhang2025prlm,zhang2025reward}. However, the prevailing paradigm in these studies follows a \emph{reasoning-first} structure, where the rationale (e.g., wrapped in <think> tags) strictly precedes the final answer. While beneficial for accuracy, this sequential dependency imposes a severe penalty on inference latency. In real-time scenarios like query correction, the latency introduced by pre-generation reasoning is operationally unacceptable. This creates a critical dilemma: standard CoT is too slow for search, while direct generation lacks the depth for complex correction. 
Therefore, this paper explores the SandwichR method, which effectively mitigates the latency-accuracy trade-off in complex query correction.

\section{Problem Formulation and Data Construction}

\subsection{Problem Formulation}
In this paper, we focus on the task of Query Correction (QC). Formally, we are given an annotated dataset $\mathcal{D}=\{(x_i,y_i)\}_{i=1}^{N}$, where $N$ denotes the sample size.
For each sample in the dataset, $x_i$ denotes the original user query (requiring correction), and $y_i$ represents the corresponding correct query (ground truth).
While traditional QC approaches treat this as a direct generation task—mapping the original query directly to the corrected query—we aim to endow Large Language Models (LLMs) with explicit reasoning capabilities. Specifically, our objective is to generate the corrected query $y$ from the input $x$ by leveraging an intermediate reasoning process $R$.

\subsection{Data Construction}
\label{sec:data_construction}
Existing query correction benchmarks~\cite{jiang2022mcscsetspecialistannotateddatasetmedicaldomain, 10.1145/3616855.3635847, ye-etal-2025-qspell} primarily target surface-level spelling or phonetic typos while real-world search errors go beyond surface-level typos, motivating a benchmark that exercises reasoning-style correction beyond rule-based pattern matching.
Therefore we construct a dataset based on Multi-CPR~\cite{long2022multi}. We simulate real-world search errors by injecting noise into the original queries ($Q_{\mathrm{clean}}$) to generate corrupted queries ($Q_{\mathrm{noise}}$).
% \zc{
Specifically, we introduce three representative types of query errors:
\begin{itemize}
    \item \textbf{Wrong Words:} 
    Randomly substituting a word with a visually or phonetically similar, or commonly confused character to simulate spelling errors, phonetic or visual confusions.
    \item \textbf{Missing Words:} 
    Omitting a word (e.g., a function or content word) to simulate accidental deletion.
    \item \textbf{Disorder Words:} 
    Randomly swapping adjacent words
    to simulate word order errors in hurried typing or others.
\end{itemize}

\begin{figure}[t]  % 浮动位置参数：优先此处(here)/顶部(top)/底部(bottom)/单独页(page)
    \centering        % 图片居中（必须放在figure环境内）
    % 插入图片：width设置宽度（推荐用相对宽度，如0.8\textwidth），{}内是图片路径
    \includegraphics[width=0.5\textwidth]{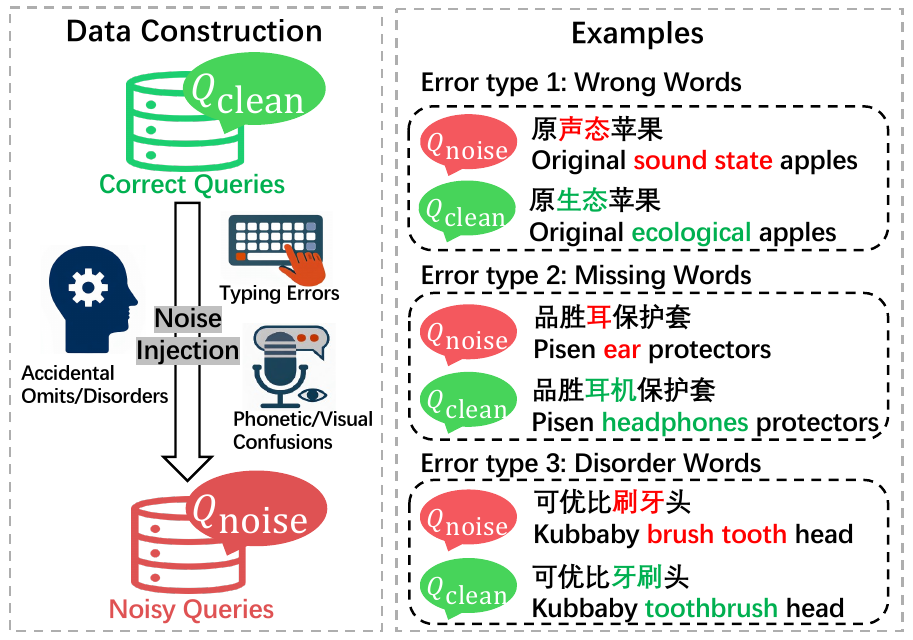}
    % 图片描述（自动在图片下方，会自动生成“图1：xxx”编号）
    \caption{Examples of three types of query errors including wrong words, missing words, disorder words.}
    \label{fig:mistake_categories}  % 可选：图片标签，用于交叉引用（如\ref{fig:labelname}）
\end{figure}

Diagrams for data construction and examples of three error types are illustrated in Figure \ref{fig:mistake_categories}. In this work, we limit each query to contain only one error. This design is based on the fact that the queries in the original dataset are relatively short, and in real-world scenarios, the error rate within such short queries is typically low, thereby making the constructed data more representative of actual search scenarios. Nevertheless, our data construction framework is flexible: higher error ratios and more complex error patterns can be readily generated by repeatedly applying or combining these three basic error types. Also, to evaluate the model's generalization capability under more complex, multi-error scenarios, we also construct an auxiliary test set characterized by high error density, where multiple overlapping errors are allowed. The detailed evaluation and robustness analysis on this high-density dataset are provided in Appendix~\ref{sec:appendix_robustness}.

The final dataset consists of pairs $(Q_{\mathrm{noise}}, Q_{\mathrm{clean}})$, also referred to as
$(x, y)$ pairs.
% serving as the ground truth for our training.

\section{The Proposed Approach: SandwichR}
In this section, we first present the overall architecture of SandwichR and then detail its two-stage training pipeline: SandwichR
Ability Acquisition via SFT and Consistency-Aware Reinforcement Learning. 
\subsection{Approach Overview} 
Let $x$ denote the input noisy query. We define the model's output $y$ as a SandwichR-structured sequence:
\begin{equation}
\label{eq:struct}
   y = [C_{\mathrm{init}}, \ R, \ C_{\mathrm{final}}],
\end{equation}
where $C_{\mathrm{init}}$ is the initial correction, $R$ represents the correction reasoning process (i.e., reasoning trajectory), and $C_{\mathrm{final}}$ is the final correction derived from the reasoning. This structure allows the model to return $C_{\mathrm{init}}$ immediately to the user, facilitating deployment in latency-sensitive applications.

\begin{figure*}[t] % 注意这里的星号 *
    \centering
    \includegraphics[width=\textwidth]{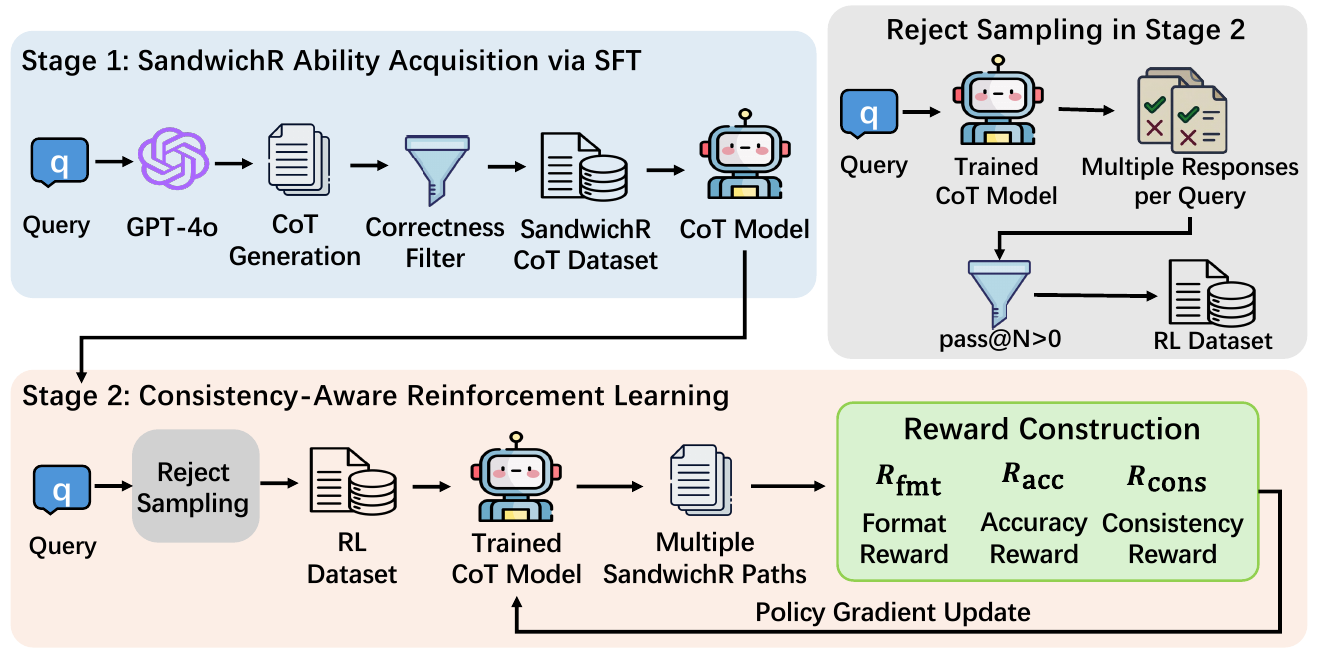}
    \caption{The overall workflow of the proposed SandwichR. It consists of two stages: (1) \textbf{SandwichR Ability Acquisition} via Supervised Fine-Tuning (SFT), and (2) \textbf{Consistency-Aware} Reinforcement Learning (RL).}
    \label{fig:overall_workflow}
\end{figure*}
Figure~\ref{fig:overall_workflow} illustrates the overall training of SandwichR.
The training of SandwichR is conducted in two stages: (1) SandwichR Ability Acquisition via Supervised Fine-Tuning~(SFT), which teaches the model to produce the SandwichR-structured output; and (2) Consistency-Aware Reinforcement Learning, which refines the model by explicitly reinforcing the alignment between the initial correction $C_{\mathrm{init}}$ and the reasoning process $R$, ensuring the first answer benefits from the subsequent reasoning for higher accuracy.

\subsection{SandwichR Ability Acquisition via SFT}
We first adapt a base LLM to the SandwichR format via Supervised Fine-Tuning.
We utilize GPT-4o to generate high-quality reasoning trajectories and corrections.

% \zc{
Specifically, we employ the \textbf{Reasoning-Answer} format (as shown in Appendix~\ref{sec:app:prompts}) to prompt GPT-4o, which instructs it to first generate its internal reasoning within <reasoning> tags, followed by the final corrected output within <answer> tags. This format aligns with the standard cognitive process of reasoning before answering, and thus ensures high-quality reasoning trajectories.

We then post-process these generations to restructure them into the SandwichR-structured sequence as Eq.~\eqref{eq:struct}. This restructured data is subsequently used for fine-tuning the base model to acquire the Sandwich Reasoning abilities.

% For completeness, all relevant prompts are provided in Appendix~\ref{sec:app:prompts}, including those for guiding GPT-4o to generate reasoning data, as well as those used to train our model to produce SandwichR and other structured outputs.
% }
% \begin{figure}[htbp]  % 浮动位置参数：优先此处(here)/顶部(top)/底部(bottom)/单独页(page)
%     \centering        % 图片居中（必须放在figure环境内）
%   \includegraphics[width=0.50\textwidth]{figs/prompt_types.jpg}
%     \caption{Comparison of Three Prompt Templates for Answer-Reasoning, Reasoning-Answer and Answer-Reasoning-Answer.}
%     \label{fig:three_types_prompts}  
% \end{figure}

\subsection{Consistency-Aware Reinforcement Learning}
\label{sec:RL}
% SFT alone is insufficient to align $C_{\mathrm{init}}$ with the reasoning-enhanced $C_{\mathrm{final}}$. We employ GRPO~\cite{shao2024deepseekmath} to further optimize the model.
% SFT alone fails to align $C_{\mathrm{init}}$ with the reasoning-enhanced $C_{\mathrm{final}}$. We employ GRPO~\cite{shao2024deepseekmath} to further optimize the model.
Although SFT teaches the model to produce the desired Answer--Reasoning--Answer structure, it does not explicitly enforce \(C_{\mathrm{init}}\) to match the reasoning-refined \(C_{\mathrm{final}}\). We further employ GRPO~\cite{shao2024deepseekmath} to  optimize the model with rewards for accuracy, format compliance, and consistency between \(C_{\mathrm{init}}\) and \(C_{\mathrm{final}}\).

\subsubsection{Margin-based Rejection Sampling}
\label{sec:reject_sampling}
Training on random samples is inefficient. We observe that RL works best when learning from ``borderline'' cases. 
% \zc{
We define ``borderline cases'' as inputs where the model exhibits inconsistent performance across multiple attempts—sometimes yielding an acceptable correction and sometimes not. These cases reside at the performance margin where the model demonstrates nascent capability but requires refinement to consistently generate correct outputs. 

Thus to select these high-value training instances, we implement a margin-based rejection sampling strategy. Specifically, for a given input, we sample $N$ independent reasoning trajectories from the CoT-finetuned model from stage 1. A trajectory is deemed acceptable if the corresponding correction answer achieves an F$_{0.5}$-score > 0 in practice. The input is added to the RL training dataset only if at least one of the $N$ sampled trajectories is acceptable (pass@N > 0); otherwise, it is rejected.
% }

\subsubsection{Reward Design}
The reward function $R_{\mathrm{total}}$ is the key driver of our approach, composed of three terms:
\begin{align*}
\label{eq:reward}
   R_{\mathrm{total}} = w_{\mathrm{acc}} \cdot R_{\mathrm{acc}} + w_{\mathrm{fmt}} \cdot R_{\mathrm{fmt}} + w_{\mathrm{cons}} \cdot R_{\mathrm{cons}}
\end{align*}
\begin{itemize}
    \item Accuracy Reward ($R_{\mathrm{acc}}$): Given the high precision requirement of query correction (avoiding over-correction), we use the $F_{0.5}$ score between $C_{\mathrm{init}}$ and the ground truth, rather than simple accuracy or F1 score.
    % \zc{
    In practice, to encourage meaningful corrections, $R_{\mathrm{acc}}$ is set to $0$ if $C_{\mathrm{init}}$ is identical to the original query $Q_{\mathrm{noise}}$, and to the $F_{0.5}$ score otherwise.
    \begin{align*}
R_{\mathrm{acc}} = \begin{cases}
0,  \text{if } C_{\mathrm{init}} = Q_{\mathrm{noise}}, \\
F_{0.5}(C_{\mathrm{init}}, Q_{\mathrm{noise}},Q_{\mathrm{clean}}),  \text{otherwise.}
\end{cases}
\end{align*}
% }
    %     \begin{equation}
    % \label{eq:reward 3}
    %    R_{acc} = F_{0.5}(C_{\mathrm{final}}, Q_{\mathrm{clean}})
    % \end{equation}
    \item Format Reward ($R_{\mathrm{fmt}}$): A binary reward that penalizes the model if it fails to follow the strict [Correct -> Reason -> Correct] structure, ensuring parsability.
    \item Consistency Reward ($R_{\mathrm{cons}}$): To force the model to internalize the reasoning into the first step, we reward the identity between the initial and final output:
    % conditional on the final output being correct:
    \begin{align*}
    %\label{eq:reward 3}
       R_{\mathrm{cons}} = \mathbb{I}(C_{\mathrm{init}} = C_{\mathrm{final}}), 
       % \cdot \mathbb{I}(C_{\mathrm{final}} \text{ is correct})
    \end{align*}
    where \(\mathbb{I}(\cdot)\) is the indicator function (1 if the condition holds, 0 otherwise).  
    This reward is crucial because it discourages two undesirable behaviors: ``blind guessing'', where \(C_{\mathrm{init}}\) happens to be correct but \(C_{\mathrm{final}}\) changes it after reasoning, and ``disconnect'', where the reasoning process fixes the error in \(C_{\mathrm{final}}\) but \(C_{\mathrm{init}}\) remains incorrect. The role of \(R_{\mathrm{cons}}\) is therefore to align the initial correction with the model's reasoning-refined final correction, rather than to measure correctness directly. Correctness is captured by \(R_{\mathrm{acc}}\), while \(R_{\mathrm{cons}}\) encourages the first-pass correction to remain consistent with the model's post-reasoning decision. Together, these rewards guide the model toward initial corrections that are both accurate and self-aligned.
   % \zc{
   In practice, the format and consistency requirements are jointly enforced by computing $R_{\mathrm{fmt}} \times R_{\mathrm{cons}}$ as a unified term. This yields a reward of 1 only when the output adheres to the required structure and $C_{\mathrm{init}}$ matches $C_{\mathrm{final}}$.
    % We provide a theoretical discussion in Appendix~\ref{sec:theory}, where this reward ensures that the accuracy of $C_{\mathrm{init}}$ is close to the accuracy of $C_{\mathrm{final}}$.
   % }
\end{itemize}
\subsection{Discussion}
\label{sec:discussion}

A potential concern with the Answer--Reasoning--Answer format is the causal constraint of autoregressive decoding. Since the initial correction \(C_{\mathrm{init}}\) is generated before the reasoning tokens \(R\), it cannot condition on \(R\) within a single forward pass. SandwichR therefore should not be interpreted as making the first answer directly use future reasoning tokens.

Instead, SandwichR can be viewed as a form of \textbf{structured self-distillation}. The same policy generates an initial correction \(C_{\mathrm{init}}\), a reasoning trace \(R\), and a reasoning-refined final correction \(C_{\mathrm{final}}\). During training, \(C_{\mathrm{final}}\) acts as the model's own post-reasoning target, and the consistency-aware RL objective encourages \(C_{\mathrm{init}}\) to agree with this target. In this sense, SandwichR distills the model's post-reasoning decision into its first-pass correction through a structured output format and a sequence-level consistency reward.

We use the empirical agreement rate
\begin{align*}
    \alpha = P(C_{\mathrm{init}} = C_{\mathrm{final}})
\end{align*}
to quantify how often the low-latency initial correction matches the reasoning-refined final correction. A high value of \(\alpha\) should be interpreted as evidence of strong self-alignment rather than as a formal correctness guarantee. As shown in Sec.~5.8, consistency-aware RL yields high agreement between \(C_{\mathrm{init}}\) and \(C_{\mathrm{final}}\) across all domains.
\section{Experiments}
This section presents our experimental setup and comprehensive results, covering datasets, evaluation metrics, baseline models, implementation details, and analysis of key findings.
\subsection{Experimental Settings}
\subsubsection{Dataset and Metric}

We conducted experiments on three error correction datasets constructed based on Multi-CPR-QC, namely E-commerce, Video, and Medical.
These datasets contain queries from different domains, and all original queries were collected from real search engine systems within Alibaba Group~\cite{long2022multi}.
The statistical analysis of these three datasets is presented in Table~\ref{tab:data_main}.
Following the commonly used evaluation metrics in the error correction field~\cite{zhang2025trigger3,xu2022fcgec}, we adopted F$_{0.5}$-score and Accuracy (Acc) to evaluate the model's error correction performance.
\begin{table}[htbp]
\centering
\caption{Statistics of our datasets. \textbf{\#Train}, \textbf{\#Dev}, \textbf{\#Test} denote the number of the train, development and test datasets, while \textbf{Tra\_Len}, \textbf{Dev\_Len} and \textbf{Tes\_Len} represent their average query lengths, respectively.}
\resizebox{\columnwidth}{!}{
\begin{tabular}{lcccccc}
    \toprule
    \textbf{Dataset} & \textbf{\#Train} & \textbf{\#Dev} & \textbf{\#Test} & \textbf{Tra\_Len} & \textbf{Dev\_Len} & \textbf{Tes\_Len} \\
    \midrule
    \textbf{E-commerce} & 90,511 & 1,000 & 1,000 & 6.43 & 6.39 & 6.45 \\
    \textbf{Video}      & 88,736 & 1,000 & 1,000 & 7.09 & 7.14 & 7.12 \\
    \textbf{Medical}    & 94,176 & 1,000 & 1,000 & 16.08 & 16.19 & 16.34 \\
    \bottomrule
\end{tabular}
}
\label{tab:data_main}
\end{table}
\subsubsection{Data Selection and Reuse Protocol}
\label{sec:data_protocol}
To ensure a rigorous and fair comparison across all baselines, we establish a strict protocol for training data selection and reuse from the larger pool of 89k--94k queries per domain. For Supervised Fine-Tuning, we apply the dual-filtering process described in Sec.~\ref{sec:data_construction} (retaining only correct reasoning-correction pairs from GPT-4o) and randomly sample a subset of 1,000 queries per domain, ensuring an even distribution across the Wrong, Missing, and Disorder error types. Crucially, to guarantee absolute fairness, this identical set of 1,000 high-quality queries is adapted into the respective formats required by each baseline (i.e., \textit{Rea-Ans}, \textit{Ans-Rea}, and our \textit{SandwichR}) and reused across all SFT experiments. For the reinforcement learning stage, we draw 200 samples from the remaining training pool using the margin-based rejection sampling detailed in Sec.~\ref{sec:reject_sampling} (pass@$N > 0$, $N=4$). Since this RL filtering process inherently depends on the generation capability of the model itself, the sampling is performed independently using each method's respective SFT checkpoint, resulting in method-specific RL training sets. All hyperparameter tuning decisions are strictly based on a separate 1,000-sample development set.

\subsubsection{Baselines}
For the baselines, we selected models from different domains for testing, including \textbf{mT5}~\cite{xue2021mt5massivelymultilingualpretrained} and \textbf{BART}~\cite{lewis2019bartdenoisingsequencetosequencepretraining}—two representative models in the traditional error correction field.
% We conducted tests on the selected base models under two settings:
% \textbf{Zero-shot}: The model was directly tasked with error correction without any prior examples.
% \textbf{Few-shot}: The model was guided to perform error correction with several annotated examples.
We adopted three Chain-of-Thought (CoT) prompting strategies (see Appendix ~\ref{sec:app:prompts} for their full prompts):
\textbf{Ans-Rea}: The model follows an Answer-Reasoning format, presenting the correction outcome first, followed by the reasoning process.
\textbf{Rea-Ans}: The model follows a Reasoning-Answer format, providing the reasoning process prior to the final correction result.
\textbf{SandwichR}: The model uses an Answer-Reasoning-Answer format for output.
Additionally, we implemented two optimization methods:
\textbf{x-SFT}: Enhancing the model’s correction capability via supervised fine-tuning.
\textbf{x-RL}: Further boosting performance by integrating reinforcement learning.
% In Section~\ref{sec:}, we explored replacing the base model with DeepSeek-R1-Distill-Qwen-1.5B. While this model has been distilled on reasoning datasets, its design primarily focuses on mathematical reasoning tasks.
\textbf{GrammarGPT-7B}~\cite{fan2023grammargpt}: A grammar error correction LLM that has been SFT on a grammar correction dataset.
We also evaluated larger-scale models, including DeepSeek-R1-Distill-Qwen-7B (\textbf{Deepseek-R1-7B}), DeepSeek-R1-Distill-Qwen-32B (\textbf{Deepseek-R1-32B}), \textbf{QwQ-32B-Preview}~\cite{qwq-32b-preview},\textbf{GPT-4o-mini}, \textbf{GPT-4o},  and \textbf{DeepSeek-R1}.

\begin{table*}[htbp]
    \small
\caption{Performance comparison between SandwichR and the baseline on the three datasets, with the best performance among the trained LLMs is highlighted in bold, and the second-best method is underlined.}
    \label{tab:main exp}
\centering
    \begin{tabular}{lc cc cc cc}    
        \toprule &
        &\multicolumn{2}{c}{\textbf{E-commerce}}
        &\multicolumn{2}{c}{\textbf{Video}}
        &\multicolumn{2}{c}{\textbf{Medical}}
        \\
         \textbf{Category} &\textbf{Models}  & \textbf{F$_{0.5}$} & \textbf{Acc} & \textbf{F$_{0.5}$} & \textbf{Acc} & \textbf{F$_{0.5}$} & \textbf{Acc}\\
        \midrule
        \multirow{2}{*}{Traditional Model}  
        &mT5 SFT & 0.09 & 0.079 & 0.198 & 0.165 & 0.166 & 0.117 \\
        &BART SFT & 0.150 & 0.124 & 0.303 & 0.27 & 0.377 & 0.35 \\
        \midrule
        % \multirow{5}{*}{Reasoning LLM w/o Training}  
        % &Zero-Shot   & 0.139 & 0.110 & 0.138 & 0.091 & 0.135 & 0.049 \\
        % &Few-shot  & 0.135 & 0.108 & 0.134 & 0.084 & 0.144 & 0.055 \\
        % &Rea-Ans & 0.059 & 0.041 & 0.059 & 0.034 & 0.083 & 0.019 \\
        % &Ans-Rea  & 0.113 & 0.094 & 0.100 & 0.068 & 0.123 & 0.044 \\
        % &SandWichR & 0.089 & 0.068 & 0.095 & 0.063 & 0.112 & 0.045 \\
        % \cdashline{2-8}
        % \multirow{6}{*}{Base LLM w/ Training \\ (1.5B scale)}
        \multirow{6}{*}{Trained Base LLM (1.5B)}
        % {\parbox{3cm}{\centering Base LLM w/ Training \\ (1.5B scale)}}
        &Rea-Ans SFT & 0.178 & 0.164 & 0.278 & 0.255 & 0.347 & 0.309\\
        &Ans-Rea SFT  & 0.199 & 0.181 & 0.280 & 0.250 & 0.366 & 0.312\\
        &SandwichR SFT & 0.202 & 0.187 & 0.293 & 0.253 & 0.376 & 0.338\\
        &Ans-Rea SFT+RL  & 0.211 & 0.200 & 0.316 & 0.292 & \underline{0.392} & 0.363\\
        &Rea-Ans SFT+RL & \underline{0.216} & \underline{0.207} & \underline{0.318} & \underline{0.301} & 0.387 & \underline{0.364}\\
        % &SandWichR SFT+RL(Ours) & 0.221 & 0.213 & 0.325 & 0.307 & 0.396 & 0.375\\
        &SandwichR SFT+RL (Ours) & \textbf{0.221} & \textbf{0.213} & \textbf{0.325} & \textbf{0.307} & \textbf{0.396} & \textbf{0.375}\\
        \midrule
        \multirow{7}{*}{Large-scale LLM}  
        &GrammarGPT-7B & 0.045 & 0.037 & 0.092 & 0.081 & 0.161 & 0.148 \\
        &Deepseek-R1-7B & 0.071 & 0.053 & 0.095 & 0.064 & 0.140 & 0.072 \\
        &Deepseek-R1-32B & 0.212 & 0.185 & 0.249 & 0.203 & 0.350 & 0.261 \\
        &GPT-4o-mini & 0.227 & 0.199 & 0.296 & 0.264 & 0.392 & 0.316 \\
        &QwQ-32B-Preview & 0.295 & 0.214 & 0.329 & 0.256 & 0.427 & 0.301 \\
        &GPT-4o & 0.299 & 0.259 & 0.354 & 0.310 & 0.475 & 0.384 \\
        &Deepseek-R1 & 0.333 & 0.244 & 0.371 & 0.284 & 0.452 & 0.321 \\
        \bottomrule
    \end{tabular}
\end{table*}

\subsubsection{Implementation Details}
% All our training and testing were conducted on GPUxx.
% During GRPO training, we set the learning rate to 1e-6, the number of generations (num\_generations) to 4, and the maximum completion length to 768 or 1024 tokens. We adopted LoRA for efficient fine-tuning of the large language model, with LoRA hyperparameters set as follows: r = 16, alpha = 16.The values of $\alpha$ and $\beta$ were both set to 0.1.
% For model training, we leveraged DeepSpeed ZeRO-2~\cite{rajbhandari2020zero} for distributed optimization, and we employed vLLM~\cite{kwon2023efficient} to accelerate the inference process.

% \zc{
We adopt Qwen2.5-1.5B-Instruct as our base LLM. This relatively small-scale model is chosen for its suitability to query correction where low latency and deployment cost are critical.

All our training and testing were conducted on NVIDIA A100 40GB GPUs. 
During SFT training, we performed LoRA~\cite{hu2021loralowrankadaptationlarge} fine-tuning. The LoRA hyperparameters were set to r=8 and alpha=16. 
Training was performed on a curated dataset of 1,000 samples with a batch size of 32 and a learning rate of 5e-5. 
% The model was trained for 30 epochs, and the checkpoint that achieved the best performance on the validation set was selected. 
% This optimal checkpoint served both as the final model for the SFT stage evaluation and as the initial policy model for the subsequent GRPO training. 
During GRPO training, we conducted full-parameter fine-tuning which utilized a dataset of 200 samples with a batch size of 8. 
We employed a learning rate of 1e-5 and 20 epochs.
% with a cosine decay schedule, which decreased to a final value of 1e-6. 
The maximum completion length was 256 tokens.
% Models were trained for 20 epochs.
% and the checkpoint that performed best on the validation set was selected for final evaluation.
More experimental details
can be found in Appendix~\ref{sec: implementation details} and~\textcolor{blue}{\url{https://github.com/zcbalak/SandwichR}}.
% }

\subsection{Main Results}

\begin{table*}[t]
\caption{Robustness under Strict Token Budgets. Time in seconds per sample. Full budget allows 256 tokens; Limited budget simulates low-latency constraints. $\Delta$ indicates the change from Full to Limited Budget.}
% The cross symbol (†) marks a catastrophic and undesirable drop in accuracy.}
\centering
\small
\resizebox{0.85\textwidth}{!}{ % 宽度可自行调整
\begin{tabular}{l r ccc ccc ccc}
\toprule
\multirow{2}{*}{Method} & \multirow{2}{*}{Setting} & \multicolumn{2}{c}{E-commerce} & \multicolumn{2}{c}{Video} & \multicolumn{2}{c}{Medical} \\
\cmidrule(lr){3-4} \cmidrule(lr){5-6} \cmidrule(lr){7-8}
 & & Acc & Time (s) & Acc & Time (s) & Acc & Time (s) \\
\midrule
\multirow{3}{*}{Rea-Ans}
    & Full Budget & 0.207 & 1.959 & 0.301 & 1.143 & 0.364 & 1.550 \\
    & Limited Budget & 0.000 & 0.457 & 0.000 & 0.484 & 0.009 & 0.900 \\
    & $\Delta$ (\%) & 
    \textbf{-100.00} & -76.67 & \textbf{-100.00} & -57.66 & \textbf{-97.53} & -41.94 \\ %$^{\dagger}$
\midrule
\multirow{3}{*}{Ans-Rea }
    & Full Budget & 0.200 & 1.487 & 0.292 & 2.182 & 0.363 & 1.089 \\
    & Limited Budget & 0.200 & 0.464 & 0.292 & 0.446 & 0.359 & 0.893 \\
    & $\Delta$ (\%) & \textbf{0.00} & -68.80 &\textbf{ 0.00} & -79.56 & \textbf{-1.10} & -18.00 \\
\midrule
\multirow{3}{*}{SandwichR}
    & Full Budget & 0.213 & 1.613 & 0.307 & 1.174 & 0.375 & 1.683 \\
    & Limited Budget & 0.213 & 0.467 & 0.307 & 0.474 & 0.374 & 0.924 \\
    & $\Delta$ (\%) & \textbf{0.00} & -71.05 & \textbf{0.00} & -59.63 & \textbf{-0.27} & -45.10 \\
\bottomrule
\end{tabular}
} % 注意：此处是 \resizebox 的结束括号
\label{tab:efficiency}
\end{table*}
\begin{table*}[t]
    \small
    \caption{Analysis of accuracy of Different Error Types for SFT+RL Models on Three Datasets.Best performance among each error type and dataset is highlighted in bold.}
    \label{tab:error_analysis_sft_rl}
    \centering
    \resizebox{0.7\textwidth}{!}{
        \begin{tabular}{lrccc}    
            \toprule
            \textbf{Dataset} & \textbf{Model} & \textbf{Wrong Words} & \textbf{Missing Words} & \textbf{Disorder Words} \\
            \midrule
            \multirow{3}{*}{Medical} 
            & Ans-Rea & 0.3904 & 0.3003 & 0.3982 \\
            & Rea-Ans & 0.3694 & 0.3003 & \textbf{0.4222} \\
            & SandwichR & \textbf{0.3934} & \textbf{0.3153} & 0.4162 \\
            \midrule
            \multirow{3}{*}{E-commerce} 
            & Ans-Rea & 0.1862 & 0.1317 & 0.2823 \\
            & Rea-Ans & 0.1532 & 0.1407 & \textbf{0.3273} \\
            & SandwichR & \textbf{0.1892} & \textbf{0.1467} & 0.3033 \\
            \midrule
            \multirow{3}{*}{Video} 
            & Ans-Rea & 0.2733 & 0.2312 & 0.3713 \\
            & Rea-Ans & 0.2853 & \textbf{0.2492} & 0.3683 \\
            & SandwichR & \textbf{0.2913} & 0.2462 & \textbf{0.3832} \\
            \midrule
            Average & \textbf{--} & 0.2813 & 0.2291 & 0.3636 \\
            \bottomrule
        \end{tabular}
    }
    % \vspace{-3mm}
\end{table*}

We evaluate the performance of our proposed method, SandwichR, against various baselines across three query correction datasets. The main results are presented in Table~\ref{tab:main exp}. Based on the experimental outcomes, we draw the following conclusions:

\textbf{SandwichR achieves superior performance.} As shown in Table~\ref{tab:main exp}, our complete method SandwichR, refined through both SFT and RL, consistently achieves the best correction performance across all three datasets among models of similar scale, demonstrating the effectiveness of our approach. It outperforms all traditional correction models and even surpasses some untrained, larger-scale models (e.g., DeepSeek-R1-32B) in correction performance, demonstrating the efficacy of our two-stage post-training pipeline.
Traditional Seq2Seq models exhibit lower performance due to their lack of the deep semantic understanding and reasoning capabilities inherent in LLMs. Some of the untrained large-scale LLMs, despite their vast general knowledge, show suboptimal performance due to the lack of task-specific adaptation.
While some larger-scale LLMs (e.g., GPT-4o) achieve higher accuracy, they come with prohibitive computational cost and latency, making them impractical for real-time search. SandwichR offers the optimal balance for deployment-sensitive scenarios.

\textbf{Effectiveness of the SandwichR Architecture.} To isolate the effect of the architecture, we compare models built on the same base LLM with different reasoning formats. Post-training analysis reveals that the SandwichR strategy consistently outperforms both Rea-Ans and Ans-Rea structures across all datasets, confirming the validity of this structural design. By leveraging the autoregressive nature of LLMs, this structure ensures that the reasoning process explicitly informs and refines the final result ($C_{\mathrm{final}}$).Crucially, the enforced consistency between the initial ($C_{\mathrm{init}}$) and final results allows us to achieve high efficiency (by utilizing $C_{\mathrm{init}}$ for inference) without compromising quality. The SandwichR structure effectively guides the model to internalize the reasoning process required for complex query correction.

\textbf{RL further enhances model performance.} 
% Comparing trained and untrained models, it is evident that training significantly improves correction accuracy.
While Supervised Fine-Tuning (SFT) improves performance to a certain extent, we observe that Reinforcement Learning (RL) consistently outperforms SFT across all three datasets.
This indicates that RL can further optimize the model's reasoning preferences and alignment following SFT. It is particularly effective for training LLMs to possess robust reasoning capabilities for query correction tasks.
\subsection{Efficiency Analysis}
\label{sec:efficiency}
% \zc{
To evaluate the practical efficiency of different reasoning formats, we conducted a comparative analysis of three reasoning formats: Rea-Ans, Ans-Rea, and SandwichR—under two distinct token budget conditions: a sufficient budget (256 tokens) and a strictly limited budget (20 tokens for E-commerce and Video queries; 40 tokens for the longer Medical queries) to simulate latency-sensitive deployment scenarios. The results, including inference latency and accuracy, are presented in Table \ref{tab:efficiency}.

Under token limits, both Ans-Rea and SandwichR significantly reduce inference time with only a minor accuracy drop, since both formats allow the corrected query to be returned before generating reasoning tokens. However, SandwichR consistently achieves higher accuracy under the same low-latency setting. While Ans-Rea generates reasoning only after the early correction, SandwichR further introduces a reasoning-refined final correction during training and aligns the early correction with it. This self-alignment signal helps the initial correction approximate the model's post-reasoning decision, improving accuracy without increasing user-facing latency. In contrast, the widely-adopted standard CoT mode, Rea-Ans, often fails to output a final answer before reaching the token limit, as its initial reasoning consumes available tokens. SandwichR overcomes this limitation and achieves 40\%--70\% relative inference speedup compared to Rea-Ans while maintaining comparable accuracy. While absolute latency does not yet meet real-time search engine requirements, SandwichR narrows this gap algorithmically, validating its structural advantage.

% }

% \subsection{Cross-Domain Generalization}

% \subsection{Different Architectures}

% \subsection{Generalization on retrieval tasks}

\subsection{Analysis of Different Error Types}

% \zc{
% We analyze model performances on the three error types: Wrong Words, Missing Words, and Disorder words as illustrated in Table \ref{tab:error_analysis_sft_rl}. The overall average accuracy ranks these tasks by difficulty: Disorder words is the easiest, followed by Wrong Words , with Missing Words being the most challenging.

% Our SandwichR model demonstrates consistent advantages. It achieves the best accuracy for Missing Words—the hardest task—on both the Medical and E-commerce datasets. Furthermore, it delivers the highest accuracy for Wrong Words on the Medical dataset and for Disorder words on the Video dataset. Comparatively, SandwichR outperforms Ans-Rea across all 9 (Dataset × Error Type) categories and surpasses Rea-Ans in 6 out of 9 categories. This robust and comprehensive performance across diverse errors highlights the effectiveness of the Sandwich Reasoning paradigm in providing reliable verification and correction.

We analyze model performances on three error types: Wrong Words, Missing Words, and Disorder words as illustrated in Table \ref{tab:error_analysis_sft_rl}. The overall average accuracy ranks these tasks by difficulty, with Disorder words being the easiest, followed by Wrong Words, and Missing Words the most challenging. SandwichR shows consistent advantages: it achieves the best accuracy for the hardest task Missing Words on both Medical and E-commerce datasets, for Wrong Words on Medical, and for Disorder words on Video. It outperforms Ans-Rea in all 9 (Dataset × Error Type) categories and surpasses Rea-Ans in 6 out of 9.
% demonstrating the robustness of the sandwich reasoning paradigm.
This robust and comprehensive performance across diverse errors highlights the effectiveness of the Sandwich Reasoning paradigm in providing reliable correction.
% }

We also conducted multi-task merged training (details in Appendix~\ref{sec:appendix task merge}), which yielded better performance by enabling the model to learn generalizable features from cross-task data.

% \begin{figure}[htbp]  % 浮动位置参数：优先此处(here)/顶部(top)/底部(bottom)/单独页(page)
%     \centering        % 图片居中（必须放在figure环境内）
%     % 插入图片：width设置宽度（推荐用相对宽度，如0.8\textwidth），{}内是图片路径
%     \includegraphics[width=0.5\textwidth]{figs/error_type_analysis.png}
%     \caption{Analysis of accuracy of Different Error Types for SFT+RL Models on Three Datasets}
%     \label{fig:error_analysis_sft_rl}
% \end{figure}
\begin{table}[htbp]
\caption{Downstream retrieval Recall@1000 of SFT+RL Models across three domains}
\label{tab:retrieval_top100}
\centering
\small
\begin{tabular}{lccc}
\toprule
\textbf{Method} & \textbf{E-commerce} & \textbf{Video} & \textbf{Medical} \\
\midrule
Ans-Rea & 0.5980 & 0.5000 & 0.5560 \\
Rea-Ans & 0.6010 & 0.4990 & 0.5520 \\
\midrule
\textbf{SandwichR} & \textbf{0.6150} & \textbf{0.5180} & \textbf{0.5630} \\
\bottomrule
\end{tabular}
\end{table}

\subsection{Evaluation on Downstream Task}
\label{sec:downstream_evaluation}

Since our dataset is constructed based on Multi-CPR\cite{long2022multi}, which provides both the ground-truth query-to-document pairs and the official candidate document corpora for each domain, we can directly evaluate the practical impact of query correction on downstream search performance. We feed the corrected queries generated by the fully-trained (SFT+RL) variants into the standard Multi-CPR retrieval pipeline to search against the respective domain-specific corpora, and report Recall@1000 as the evaluation metric.

As shown in Table~\ref{tab:retrieval_top100}, SandwichR consistently achieves the highest recall across all three domains, outperforming both Rea-Ans and Ans-Rea by $1.1$--$1.9$ absolute points.  These results validate that by aligning the fast initial answer $C_{\text{init}}$ with the reasoning-guided final correction, SandwichR successfully translates intrinsic correction quality into downstream search utility, while preserving its low-latency advantage.

\subsection{Ablation Study}
\label{sec:ablation_study}

The ablation study in Table~\ref{tab:ablation} isolates the contribution of each training component across all three domains. Comparing \textit{SandwichR SFT} with the full \textit{SFT+RL} configuration, the consistency-aware RL stage yields consistent performance gains across all datasets. While SFT teaches the model to output the SandwichR structure, it cannot enforce semantic alignment between $C_{\text{init}}$ and $C_{\text{final}}$; the consistency-aware RL explicitly bridges this gap, successfully channeling reasoning-conditioned correctness into the first-pass answer. Furthermore, removing the rejection sampling strategy and training RL on randomly sampled queries consistently degrades accuracy, validating that ``borderline'' instances carry far more valuable learning signals than others. Overall, both the consistency-aware RL and the margin-based rejection sampling are individually necessary, and their combination is what enables SandwichR to achieve optimal accuracy.

\begin{table}[htbp]
\caption{Ablation Study on Training Components}
\label{tab:ablation}
\centering
\small
\resizebox{\columnwidth}{!}{
\begin{tabular}{l cc cc cc}
\toprule
\multirow{2}{*}{\textbf{Models}} & \multicolumn{2}{c}{\textbf{E-commerce}} & \multicolumn{2}{c}{\textbf{Video}} & \multicolumn{2}{c}{\textbf{Medical}} \\
\cmidrule(lr){2-3}\cmidrule(lr){4-5}\cmidrule(lr){6-7}
 & \textbf{F$_{0.5}$} & \textbf{Acc} & \textbf{F$_{0.5}$} & \textbf{Acc} & \textbf{F$_{0.5}$} & \textbf{Acc} \\
\midrule
SandwichR SFT+RL & \textbf{0.221} & \textbf{0.213} & \textbf{0.325} & \textbf{0.307} & \textbf{0.396} & \textbf{0.375}\\
 \hspace{1em} w/o Reject Sampling & 0.217 & 0.196 & 0.313 & 0.293 & 0.383 & 0.351 \\
\midrule
SandwichR SFT & 0.202 & 0.187 & 0.293 & 0.253 & 0.376 & 0.338\\
\bottomrule
\end{tabular}
}
\end{table}

% \subsection{Data Efficiency of RL Training}
% % \zc{
% To verify the data efficiency of the RL training stage, we scaled the training data on the E-commerce dataset. As shown in Figure~\ref{fig:rl_data_size}, increasing the data size from 200 to 500 and even 1000 samples did not lead to significant performance gains. 
% This indicates that RL training combined with rejection sampling is highly data-efficient, allowing the model to converge quickly by learning from a small set of high-quality ``borderline'' examples. Consequently, we selected the configuration of 200 samples as it offers an optimal balance between performance and computational cost.

\subsection{Data Efficiency of RL Training}

\begin{figure}[h]  % 浮动位置参数：优先此处(here)/顶部(top)/底部(bottom)/单独页(page)
    \centering        % 图片居中（必须放在figure环境内）
    % 插入图片：width设置宽度（推荐用相对宽度，如0.8\textwidth），{}内是图片路径
    \includegraphics[width=0.5\textwidth]{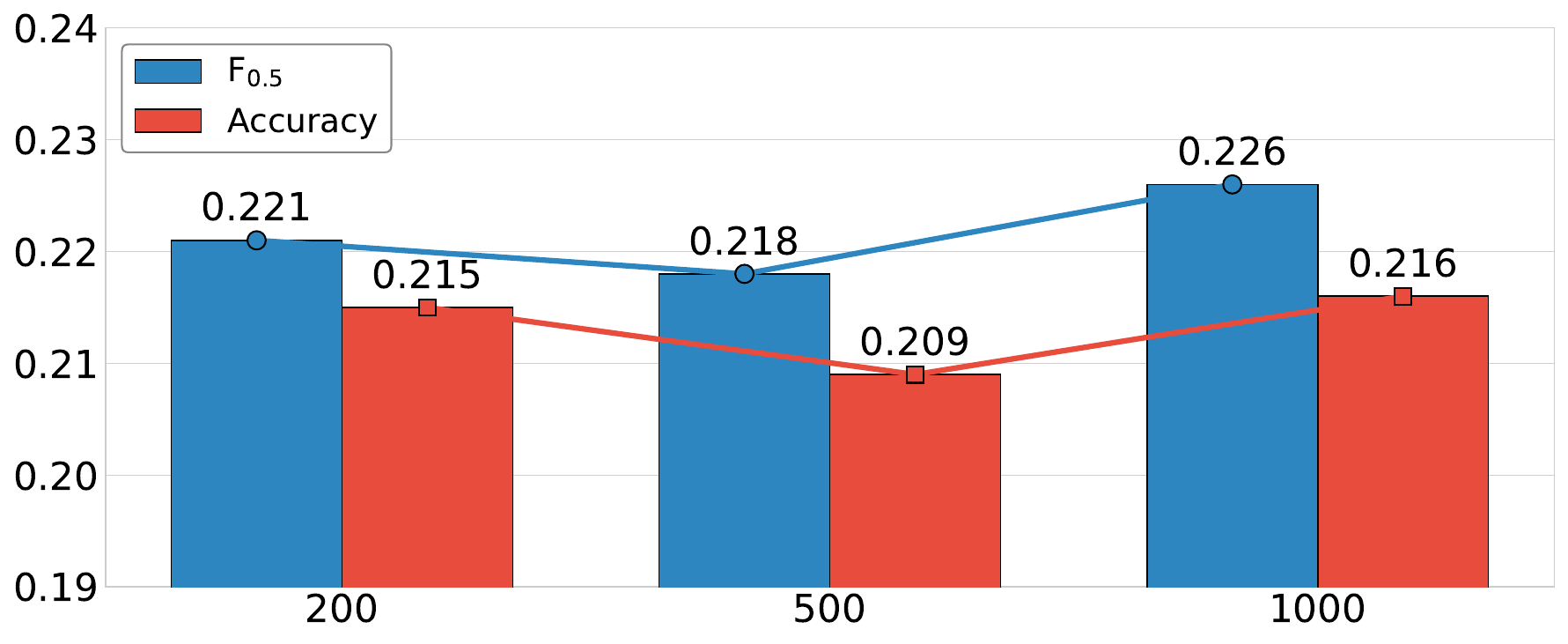}
    \caption{Performance Comparison of SandwichR with different RL training data sizes on E-commerce Dataset. The X-axis represents the RL training data size.}
    \label{fig:rl_data_size}
    % \vspace{-3mm}
\end{figure}

To verify the data efficiency of the RL training stage, we scaled the training data on the E-commerce dataset. As shown in Figure~\ref{fig:rl_data_size}, increasing the data size from 200 to 500 and even 1000 samples did not lead to significant performance gains. This suggests that the effectiveness of RL depends more on the informativeness of the selected samples than on the raw number of training instances. Since margin-based rejection sampling focuses on borderline cases where the model already shows partial correction ability, these samples provide direct learning signals for improving the first-pass correction.

Therefore, RL training combined with rejection sampling is highly data-efficient: the model can benefit from a small set of high-quality examples without requiring large-scale RL data. We select the configuration of 200 samples as it offers an effective balance between performance and computational cost.
\subsection{\mbox{Analysis on Consistency and Error Correction}}
\label{sec:consistency_analysis}

To examine whether the consistency reward indeed aligns the initial and final corrections, we conduct an agreement analysis on the SFT+RL checkpoints using 1,000 test samples per domain. We also inspect the rare disagreement cases to understand whether the final correction merely copies the initial output or can still revise residual errors after reasoning.

As shown in Table~\ref{tab:agreement}, the empirical agreement rate \(\alpha=P(C_{\mathrm{init}}=C_{\mathrm{final}})\) reaches 98.4\%--99.6\% across the three domains. This high agreement provides empirical support for the structured self-distillation interpretation in Sec.~\ref{sec:discussion}: the first-pass correction is strongly aligned with the model's reasoning-refined final correction.

\begin{table}[htbp]
\centering
\caption{Agreement analysis on SFT+RL checkpoints (1,000 test samples per domain). \(\alpha=P(C_{\mathrm{init}}=C_{\mathrm{final}})\) denotes the empirical agreement rate between the initial and final corrections.}
\label{tab:agreement}
\resizebox{\columnwidth}{!}{
\begin{tabular}{lccccc}
\toprule
\textbf{Domain} & \textbf{Total} & \textbf{Valid Format} & \(C_{\mathrm{init}} = C_{\mathrm{final}}\) & \(C_{\mathrm{init}} \neq C_{\mathrm{final}}\) & \textbf{Agreement \(\alpha\)} \\
\midrule
Medical & 1,000 & 997 & 993 & 4 & \textbf{0.9960} \\
E-commerce & 1,000 & 1,000 & 984 & 16 & \textbf{0.9840} \\
Video & 1,000 & 1,000 & 994 & 6 & \textbf{0.9940} \\
\bottomrule
\end{tabular}
}
\end{table}

The few disagreement cases further show that the final correction is not a blind copy of the initial correction. For example, in one medical query, the input is ``\begin{CJK*}{UTF8}{gbsn}看医牛\end{CJK*}'', where ``\begin{CJK*}{UTF8}{gbsn}医牛\end{CJK*}'' is a typo for ``\begin{CJK*}{UTF8}{gbsn}医生\end{CJK*}'' (doctor). The initial correction \(C_{\mathrm{init}}\) preserves this typo, while the reasoning trace diagnoses the error: \textit{``\begin{CJK*}{UTF8}{gbsn}医牛\end{CJK*} (doctor-ox) is a typo for \begin{CJK*}{UTF8}{gbsn}医生\end{CJK*} (doctor). I should correct it.''} The final correction \(C_{\mathrm{final}}\) then outputs the corrected query with ``\begin{CJK*}{UTF8}{gbsn}医生\end{CJK*}''. This example suggests that the reasoning-refined final correction can still revise residual errors in \(C_{\mathrm{init}}\), while the high agreement rate indicates that such post-reasoning decisions are largely distilled into the first-pass correction after consistency-aware RL.

Appendix~\ref{sec:distillation_advantage} further compares SandwichR with answer-only SFT and teacher-distillation baselines, highlighting the role of the structured self-distillation signal beyond answer-only supervision.

\section{Conclusion}

To bridge the gap between high accuracy and low inference latency in query correction, this paper introduces Sandwich Reasoning (SandwichR), a novel ``Answer–Reasoning–Answer'' generation framework.
It evolves from the conventional robust yet high-latency Reasoning–Answer paradigm and its low-latency but less robust variant Answer–Reasoning. 
By front-loading the answer while aligning it with subsequent reasoning, SandwichR ensures the first answer both fast and accurate, as it can benefit from the model's reasoning capability. 
We achieve this through the consistency-aware reinforcement learning strategy, which employs a consistency reward that forces the initial answer to align with the reasoning-enhanced final answer, thereby transforming the initial answer from a blind guess into a pre-emptive output that incorporates reasoning benefits.
Extensive experiments validate that SandwichR effectively balances high precision with low latency, presenting a practical solution for deployment-sensitive scenarios.
% }
\section{Limitations}
Despite its superior correction performance and high efficiency, SandwichR has several limitations that point to future research directions: \textbf{Absolute Latency vs. Production Constraints:} While SandwichR achieves a 40\%--70\% relative latency reduction compared to standard CoT, the absolute inference latency (e.g., $\sim$0.47\,s on an A100 GPU) still exceeds the ultra-strict millisecond-level requirements of some real-world search engines. Future work must combine our algorithmic framework with advanced engineering optimizations, such as model quantization and highly optimized inference engines, to fully bridge this gap; \textbf{Model Scale and External Knowledge:} Due to resource constraints, our experiments primarily utilized a 1.5B parameter LLM. Employing larger-scale LLMs with richer world knowledge is expected to yield even better correction performance. Furthermore, integrating Retrieval-Augmented Generation (RAG)~\cite{jin2025flashrag,zhang2025qe} could help resolve queries involving emerging entities or long-tail knowledge beyond the model's internal parameters.

\begin{acks}
This work was partially supported by the National Natural Science Foundation of China (No. 62376275, 62472426). Work partially done at Beijing Key Laboratory of Research on Large Models and Intelligent Governance, and Engineering Research Center of Next-Generation Intelligent Search and Recommendation, MOE. Supported by fund for building world-class universities (disciplines) of Renmin University of China.
\end{acks}

\appendix
\section{Appendix}
\subsection{Prompts}
\label{sec:app:prompts}
% \begin{figure}[htbp]  % 浮动位置参数：优先此处(here)/顶部(top)/底部(bottom)/单独页(page)
%     \centering        % 图片居中（必须放在figure环境内）
%     % 插入图片：width设置宽度（推荐用相对宽度，如0.8\textwidth），{}内是图片路径
%     \includegraphics[width=0.5\textwidth]{figs/sandwichr_Prompts.pdf}
%     % 图片描述（自动在图片下方，会自动生成“图1：xxx”编号）
%     \caption{Comparison of Reasoning Paradigms: Traditional Approach vs Our proposed SandwichR.}
%     \label{fig:inference_comparison}  % 可选：图片标签，用于交叉引用（如\ref{fig:labelname}）
% \end{figure}
This section presents the detailed prompt templates used for training and evaluation of the three reasoning formats: \textbf{Reasoning-Answer}, \textbf{Answer-Reasoning}, and our proposed \textbf{SandwichR}. Each prompt is designed to instruct the model to follow a specific output structure while performing query correction. In these templates, we use colored tags to indicate different components: the \textcolor{promptblue}{\texttt{<reasoning>...</reasoning>}} tags indicate the reasoning trace, the \textcolor{promptgreen}{\texttt{<answer>...</answer>}} tags indicate the corrected query output, and the \textcolor{promptred}{\texttt{[original query]}} placeholder represents the actual input query. The prompt structure explicitly defines the output format, ensuring the model generates responses in the desired sequence.

\begin{promptbox}{lightgray}
\textbf{Reasoning-Answer:}\\
You are a Chinese text error correction tool that can detect and correct errors in the text. Please check the errors in the following text, correct them, modify only the erroneous parts while keeping the original sentence structure as much as possible, provide your reasoning process, and output the corrected version. Please strictly use the following format for your reply: \textcolor{promptblue}{\textbf{<reasoning>}} (briefly analyze the location, type, and basis of the error) \textcolor{promptblue}{\textbf{</reasoning>}} \textbackslash n \textcolor{promptgreen}{\textbf{<answer>}} (output the corrected full text) \textcolor{promptgreen}{\textbf{</answer>}}. \textcolor{promptred}{\textbf{[original query]}}
\end{promptbox}
% \vspace{2pt}
\begin{promptbox}{lightorange}
\textbf{Answer-Reasoning:}\\
You are a Chinese text error correction tool that can detect and correct errors in the text. Please check the errors in the following text, correct them, modify only the erroneous parts while keeping the original sentence structure as much as possible, first output the corrected version, and then provide your reasoning process. Please strictly use the following format for your reply: \textcolor{promptgreen}{\textbf{<answer>}} (output the corrected full text) \textcolor{promptgreen}{\textbf{</answer>}} \textbackslash n \textcolor{promptblue}{\textbf{<reasoning>}} (briefly analyze the location, type, and basis of the error) \textcolor{promptblue}{\textbf{</reasoning>}}. \textcolor{promptred}{\textbf{[original query]}}
\end{promptbox}
% \vspace{2pt}
\begin{promptbox}{lightblue}
\textbf{SandwichR :}\\
   You are a Chinese text error correction tool that can detect and correct errors in the text. Please check the errors in the following text, correct them, modify only the erroneous parts while keeping the original sentence structure as much as possible, first output the corrected version, then provide your reasoning process, and finally output the corrected version again. Please strictly use the following format for your reply: \textcolor{promptgreen}{\textbf{<answer>}} (first output the corrected full text) \textcolor{promptgreen}{\textbf{</answer>}} \textbackslash n \textcolor{promptblue}{\textbf{<reasoning>}} (briefly analyze the location, type, and basis of the error) \textcolor{promptblue}{\textbf{</reasoning>}} \textbackslash n \textcolor{promptgreen}{\textbf{<answer>}} (output the corrected full text again) \textcolor{promptgreen}{\textbf{</answer>}}. \textcolor{promptred}{\textbf{[original query]}}
\end{promptbox}

\subsection{More Implementation Details}
\label{sec: implementation details}
During SFT training, we performed LoRA~\cite{hu2021loralowrankadaptationlarge} fine-tuning. The LoRA hyperparameters were set to r=8 and alpha=16. 
Training was performed on a curated dataset of 1,000 samples with a batch size of 32 and a learning rate of 5e-5. 
The model was trained for 30 epochs, and the checkpoint that achieved the best performance on the validation set was selected. 
This optimal checkpoint served both as the final model for the SFT stage evaluation and as the initial policy model for the subsequent GRPO training. 
During GRPO training, we conducted full-parameter fine-tuning which utilized a dataset of 200 samples with a batch size of 8. 
We employed a learning rate of 1e-5 and 20 epochs.
with a cosine decay schedule, which decreased to a final value of 1e-6. 
The maximum completion length was 256 tokens.
Models were trained for 20 epochs.
and the checkpoint that performed best on the validation set was selected for final evaluation.
For latency evaluation, latency of all models was measured on NVIDIA A100 40GB GPU using the HuggingFace Transformers library with batch size=1, greedy decoding, max\_new\_tokens=20 when limited budget or 256 when full budget, with no quantization.

\subsection{Task Merging}
\label{sec:appendix task merge}
% \zc{
We conducted a task merging experiment to evaluate the generalization capability of our SandwichR method. A single unified SandwichR model was trained on the combined datasets from all three domains (total 3,000 SFT and 600 GRPO samples), using the same training strategy, and compared against three domain-specific models trained separately on each individual dataset.

Results as shown in Table \ref{tab:mixed_training} demonstrate a clear cross-domain transfer effect: the unified model outperformed the domain-specific models on the Video and Medical datasets.
% achieving improvements of +1.8\% and +2.9\% in $F_{0.5}$ score, respectively. 
However, its performance on the E-commerce dataset was slightly lower than that of the model trained solely on E-commerce data.
% (-1.2\% in $F_{0.5}$ score). 
We attribute this pattern to two factors: (1) the inherent difficulty and lower baseline accuracy of the E-commerce task (as shown in Table 1 of the main paper), and (2) during joint training, the model likely prioritized learning patterns from the relatively simpler Video and Medical tasks, which exhibit more regular error patterns and clearer correction boundaries.
% }

\subsection{Robustness under High Error Density}
\label{sec:appendix_robustness}

As discussed in Sec.~\ref{sec:data_construction}, our primary dataset simulated realistic search queries by injecting exactly one error per query. However, real-world search environments are often unpredictable, and user queries may occasionally contain multiple, overlapping errors. To evaluate the generalization capability and robustness of SandwichR under such unseen, highly noisy environments, we conduct a zero-shot evaluation on a significantly harder E-commerce test set. This dataset features a much higher error density, allowing up to $\text{len}(q)//5$ errors per query (e.g., multiple visual, phonetic, and word-order errors within a single short query).

As shown in Table~\ref{tab:robustness}, although the absolute scores of all models drop due to the increased difficulty of the task, SandwichR still outperforms the standard reasoning-first (\textit{Rea-Ans}) paradigm by \textbf{1.2\% absolute accuracy} and the \textit{Ans-Rea} paradigm by \textbf{0.2\%}. This empirical result demonstrates that the dual-alignment mechanism of SandwichR is highly robust to severe input noise and distribution shifts. Even when trained on simpler single-error queries, the model successfully generalizes to complex, multi-error scenarios, proving its practical reliability for deployment in real-world search engines.
\begin{table}[t]
    \centering
    \caption{Performance Comparison: Single Mixed-Domain Model trained on combined data vs. Three Domain-Specific Models trained separately on each dataset using the same two-stage training strategy.}
    \label{tab:mixed_training}
    \resizebox{0.48\textwidth}{!}{ % 稍微加宽以适应更多列
    \begin{tabular}{lcc cc cc}    
        \toprule
        \multirow{2}{*}{\textbf{Training Strategy}} & 
        \multicolumn{2}{c}{\textbf{E-commerce}} & 
        \multicolumn{2}{c}{\textbf{Video}} & 
        \multicolumn{2}{c}{\textbf{Medical}} \\
        \cmidrule(lr){2-3} \cmidrule(lr){4-5} \cmidrule(lr){6-7}
        & \textbf{F$_{0.5}$} & \textbf{Acc} & \textbf{F$_{0.5}$} & \textbf{Acc} & \textbf{F$_{0.5}$} & \textbf{Acc} \\
        \midrule
        Mixed-Domain (Single Model) & 0.209 & 0.201 & \textbf{0.343} & \textbf{0.334} & \textbf{0.425} & \textbf{0.407} \\
        Domain-Specific (Three Models) & \textbf{0.221} & \textbf{0.213} & 0.325 & 0.307 & 0.396 & 0.375 \\
        \bottomrule
    \end{tabular}}
\end{table}

\begin{table}[htbp]
\centering
\caption{Zero-shot evaluation under high error density on the E-commerce dataset.}
\label{tab:robustness}
\begin{tabular}{lcc}
\toprule
\textbf{Method} & \textbf{F$_{0.5}$} & \textbf{Acc} \\
\midrule
\textit{Rea-Ans} & 0.134 & 0.102 \\
\textit{Ans-Rea} & 0.145 & 0.112 \\
\textbf{SandwichR} & \textbf{0.148} & \textbf{0.114} \\
\bottomrule
\end{tabular}
\end{table}

\subsection{Comparison with Answer-Only Baselines}
\label{sec:distillation_advantage}

SandwichR can be viewed as a structured self-distillation framework, where the reasoning-refined final correction \(C_{\mathrm{final}}\) serves as an internal teacher signal for the initial correction \(C_{\mathrm{init}}\). To examine whether this structured distillation signal is necessary, we compare SandwichR with two answer-only distillation baselines on the E-commerce dataset.

Direct SFT trains the 1.5B model directly on ground-truth corrections without reasoning traces. LLM-Teacher Distillation trains the same 1.5B model on answer-only corrections generated by Qwen2.5-7B-Instruct. Both baselines collapse the supervision into a single corrected query, whereas SandwichR preserves the intermediate reasoning process and the post-reasoning final correction as part of the training structure.

As shown in Table~\ref{tab:ablation_distill}, SandwichR outperforms both answer-only baselines. This result indicates that the benefit of SandwichR does not come merely from exposing the model to corrected answers. Instead, the structured self-distillation signal provided by \(C_{\mathrm{final}}\) helps align the low-latency initial correction with the model's post-reasoning decision, leading to stronger correction performance than answer-only supervision.
\begin{table}[htbp]
\centering
\caption{Comparison with direct answer-only SFT and teacher distillation baselines on the E-commerce dataset.}
\label{tab:ablation_distill}
\begin{tabular}{lc}
\toprule
\textbf{Method} & \textbf{F$_{0.5}$} \\
\midrule
Qwen2.5-7B $\rightarrow$ 1.5B Distillation & 0.1610 \\
Direct SFT on Ground-Truth Answers & 0.2010 \\
\textbf{SandwichR (SFT+RL)} & \textbf{0.2210} \\
\bottomrule
\end{tabular}
\end{table}

\section*{GenAI Usage Disclosure}
During the preparation of this work, we used generative AI tools to assist with code debugging and to improve the grammar and readability of the manuscript. We did not use these tools to generate data, fabricate results, or make scientific claims. All code, analyses, and manuscript text were reviewed and verified by us, and we take full responsibility for the content of this paper.
\bibliographystyle{ACM-Reference-Format}
\balance
\bibliography{sample-base}

\end{document}